\documentclass[runningheads]{llncs}

\usepackage{eccv}

\usepackage{eccvabbrv}

\usepackage{graphicx}
\usepackage{booktabs}

\usepackage[accsupp]{axessibility}  

\usepackage{hyperref}

\usepackage{orcidlink}

\usepackage{amsmath}
\usepackage{amssymb}
\usepackage{mathtools}
\usepackage{comment}
\usepackage{multirow}
\usepackage{booktabs}
\usepackage{xcolor, colortbl}
\usepackage{tabularx}
\usepackage[super]{nth}
\newcolumntype{Y}{>{\centering\arraybackslash}X}
\usepackage{enumitem}
\usepackage{algorithm}
\usepackage{algorithmic}
\usepackage{color,soul}
\usepackage{bbding}
\usepackage[capitalize,noabbrev]{cleveref}
\newcommand{\x}{{\boldsymbol x}}
\newcommand{\bb}{{\boldsymbol b}}
\DeclareMathOperator{\xw}{\Tilde{\mathbf{x}}}

\definecolor{myblue}{RGB}{66,133,244}
\definecolor{mygreen}{RGB}{51,168,83}
\definecolor{myyellow}{RGB}{251,188,3}
\definecolor{myred}{RGB}{234,67,53}
\definecolor{mygrey}{RGB}{95,99,104}
\definecolor{mypup}{RGB}{153,0,204}
\begin{document}

\title{Raindrop Clarity: A Dual-Focused Dataset for Day and Night Raindrop Removal} 

\titlerunning{Raindrop Clarity: Dual-Focused Dataset for Day Night Raindrop Removal}

\author{Yeying Jin\inst{1}\orcidlink{0000-0001-7818-9534} \and
Xin Li\inst{2}\orcidlink{0000-0002-6352-6523} \and
Jiadong Wang\inst{1}\orcidlink{0000-0001-9372-3133} \and
Yan Zhang\inst{1}\orcidlink{0000-0002-5336-7100}\and
Malu Zhang\inst{3}\orcidlink{0000-0002-2345-0974}\thanks{Corresponding author}
}

\authorrunning{Y. Jin, X. Li et al.}

\institute{National University of Singapore \and University of Science and Technology of China \and University of Electronic Science and Technology of China\\
\email{jinyeying@u.nus.edu, xin.li@ustc.edu.cn, maluzhang@uestc.edu.cn}}

\maketitle

\begin{abstract}
Existing raindrop removal datasets have two shortcomings. First, they consist of images captured by cameras with a focus on the background, leading to the presence of blurry raindrops. To our knowledge, none of these datasets include images where the focus is specifically on raindrops, which results in a blurry background. Second, these datasets predominantly consist of daytime images, thereby lacking nighttime raindrop scenarios. Consequently, algorithms trained on these datasets may struggle to perform effectively in raindrop-focused or nighttime scenarios. The absence of datasets specifically designed for raindrop-focused and nighttime raindrops constrains research in this area. In this paper, we introduce a large-scale, real-world raindrop removal dataset called Raindrop Clarity. Raindrop Clarity comprises 15,186 high-quality pairs/triplets (raindrops, blur, and background) of images with raindrops and the corresponding clear background images. There are 5,442 daytime raindrop images and 9,744 nighttime raindrop images. Specifically, the 5,442 daytime images include 3,606 raindrop- and 1,836 background-focused images. While the 9,744 nighttime images contain 4,838 raindrop- and 4,906 background-focused images. Our dataset will enable the community to explore background-focused and raindrop-focused images, including challenges unique to daytime and nighttime conditions. Our data and code are available at:~\textcolor{magenta}{\url{https://github.com/jinyeying/RaindropClarity}}
\keywords{Raindrop-focused \and Background-focused \and Nighttime}
\end{abstract}

\section{Introduction}
\label{sec:intro}
\begin{figure}[t!]
	\centering
	\captionsetup[subfig]{labelformat=empty}
	\setcounter{subfigure}{0}
	{\includegraphics[width=0.7\linewidth]{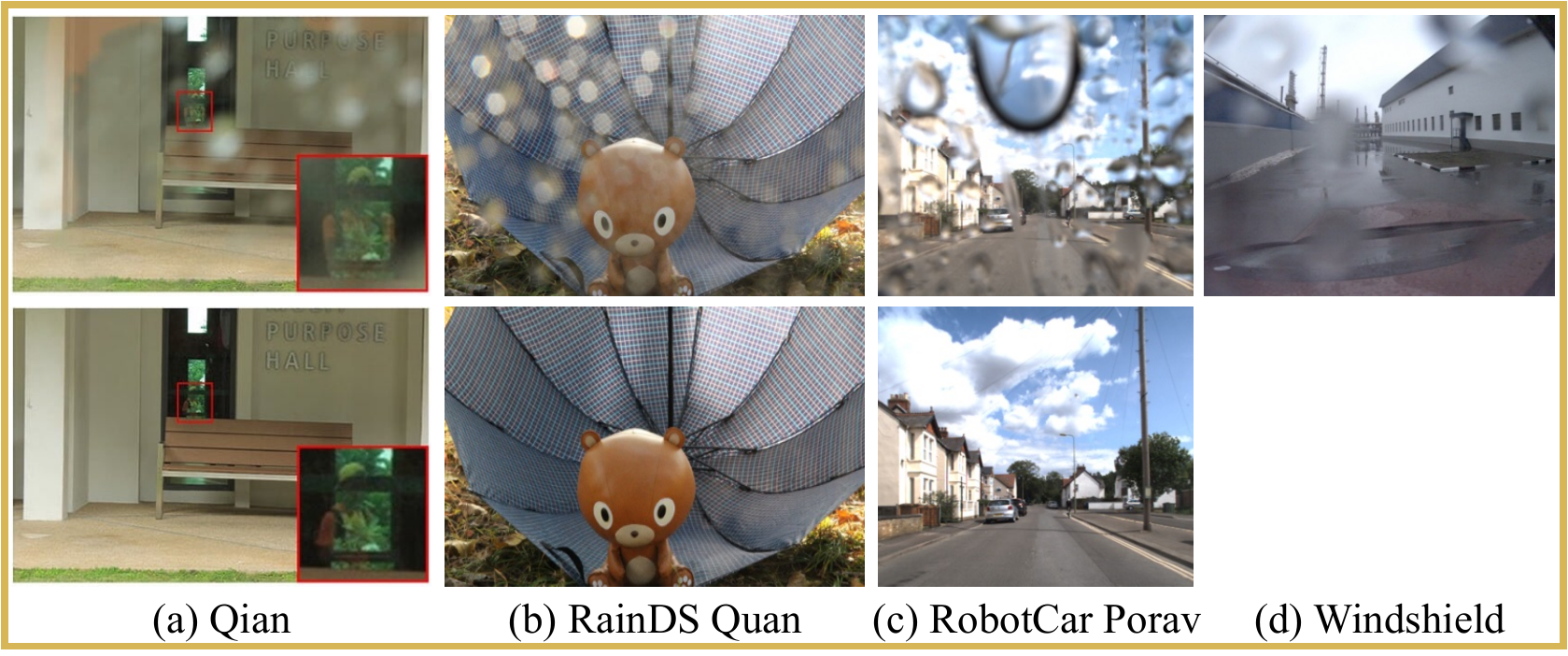}}\hfill	
	\caption{Existing raindrop datasets (e.g,~\cite{qian2018attentive,quan2021removing,porav2019can,soboleva2021raindrops}) exhibit two limitations: first, they do not include raindrop-focused images (Fig.~\ref{fig:dataset} \textcolor{myred}{right}); and second, they lack night raindrops (Fig.~\ref{fig:dataset} \textcolor{myblue}{bottom}).}
	\label{fig:existing_dataset}
\end{figure}

Adherent raindrops on lenses or windscreens can significantly reduce visibility.
Raindrop removal is crucial for surveillance, self-driving cars, object detection, outdoor photography, filmmaking, augmented reality, \etc. 
Recently, a few raindrop datasets (e.g,~\cite{qian2018attentive,quan2021removing,porav2019can,soboleva2021raindrops}) have been proposed.
In Fig.~\ref{fig:existing_dataset}, these datasets primarily contain daytime background scenes with blurry raindrops, as the images are taken with the camera's focus on the background.
They have impacts in the raindrop removal area, inspiring several successful methods~\cite{qian2018attentive,quan2019deep,porav2019can}.

\begin{figure}[t!]
	\centering
	\captionsetup[subfig]{labelformat=empty}
	\setcounter{subfigure}{0}
	{\includegraphics[width=0.73\linewidth]{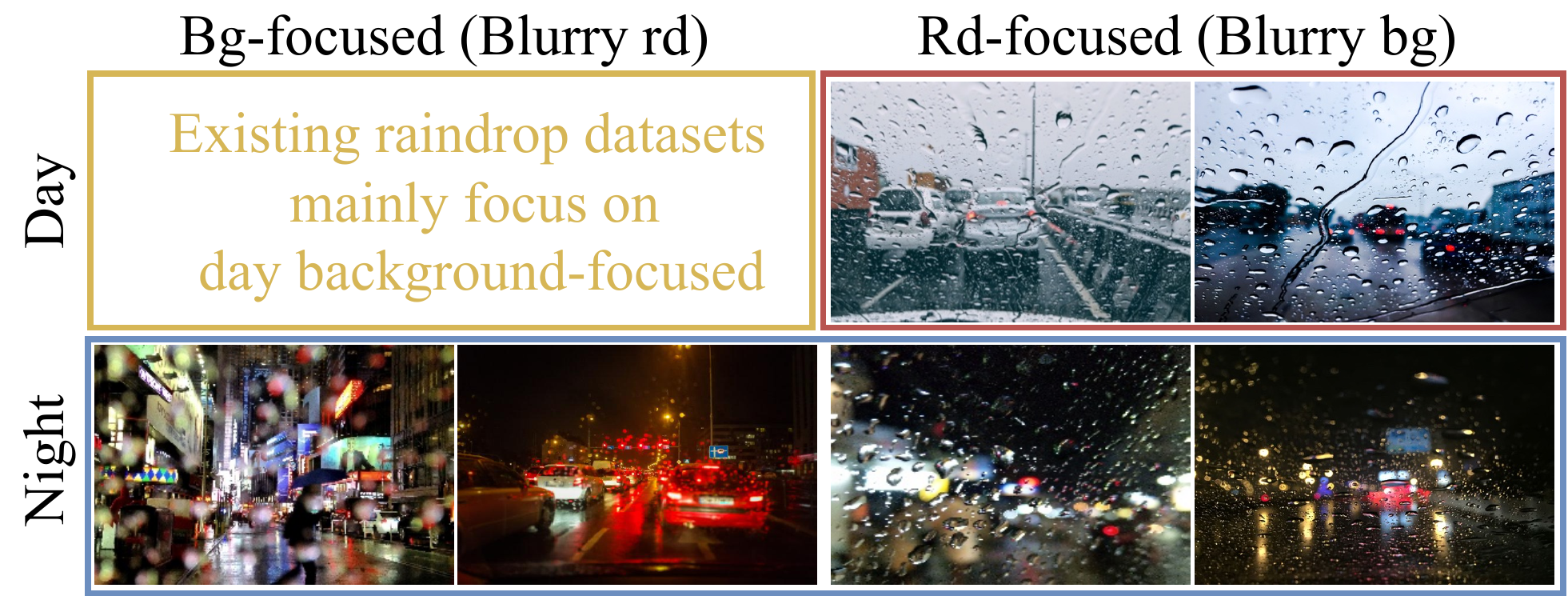}}\hfill	
	\caption{Motivation for Raindrop Clarity: real images sourced from the Internet feature scenarios overlooked by existing datasets, including raindrop-focused (\textcolor{myred}{right}) and nighttime raindrops (\textcolor{myblue}{bottom}). In this figure, Rd = Raindrop, Bg = Background.}
	\label{fig:dataset}
\end{figure}

While these datasets are valuable, they have two limitations. 
First, a significant issue arises when the camera is in automatic focus mode. 
In Fig.~\ref{fig:dataset} (\textcolor{myred}{right}), the camera may unintentionally focus on raindrops that are attached to the lens, windshield, or glass, resulting in images with sharp raindrops and a blurry background. 
Consequently, methods designed for images with blurry raindrops and a sharp background may not perform optimally.
Second, existing datasets consist only of daytime images. Therefore, existing methods, especially those reliant on training data, may struggle to effectively handle nighttime raindrop images because of the inherent domain gap between day and night conditions.

\begin{figure*}[t!]
	\centering
	\captionsetup[subfloat]{labelformat=empty}
	\setcounter{subfigure}{0}
	{\includegraphics[width=\textwidth]{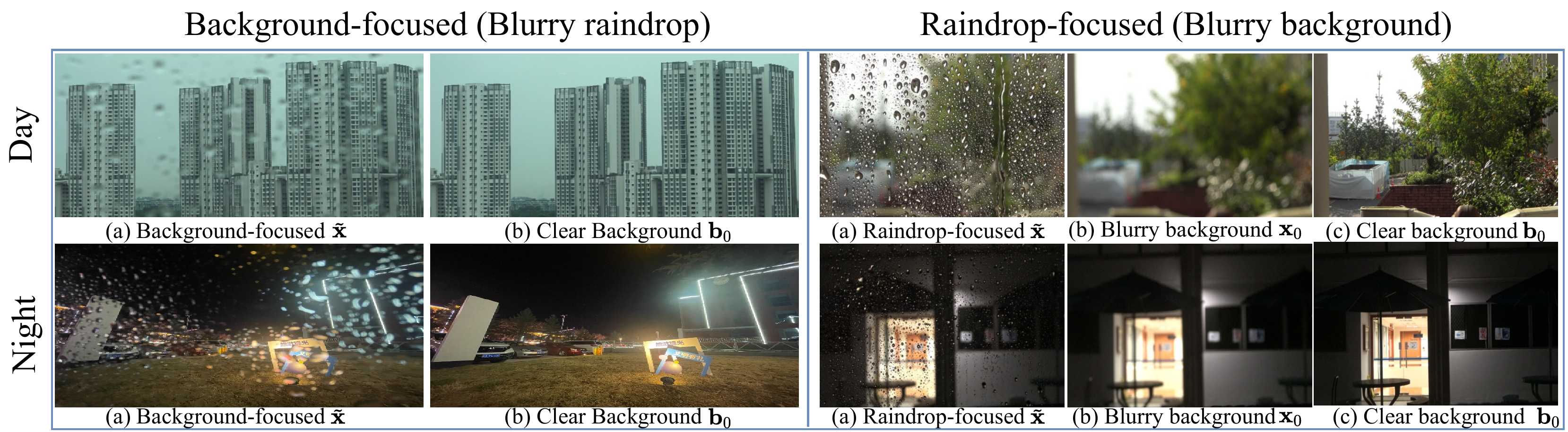}}\hfill
	\caption{Raindrop Clarity: Examples of our pairs $(\xw,\bb_0)$ and triplets $(\xw,\x_0,\bb_0)$ of a raindrop image $\xw$, the corresponding blurry background image $\x_0$ and the corresponding clear background image $\bb_0$, in day and night. For the pairs $(\xw,\bb_0)$, there is no $\x_0$, since in this case $\x_0 = \bb_0$. The reason we have the blurry background image $\x_0$ in triplets is to have the raindrop difference maps $\mathbf{\tilde{m}}$ (see Fig.~\ref{fig:mask}).}
	\label{fig:data}
\end{figure*}

In Fig.~\ref{fig:dataset}, we observe that besides daytime images with blurry raindrops and a sharp background, it is common to come across images with sharp raindrops and a blurry background (\textcolor{myred}{right}), as well as nighttime raindrops (\textcolor{myblue}{bottom}). 
Thus, we collected varied raindrop images as shown in Fig.~\ref{fig:data}:
the top-left with daytime blurry raindrops against a sharp background, the top-right with daytime sharp raindrops and a blurry background, the bottom-left with nighttime blurry raindrops and a clear background, and the bottom-right with nighttime sharp raindrops with a blurry background.

Physical solutions like rain covers or wipers can reduce the presence of raindrops, but their effectiveness is limited. In certain situations where wipers cannot be used, or wind blows rain onto the lens, these physical measures become impractical.
Physics-based methods~\cite{you2013adherent,you2015adherent,hao2019learning,Liu2019PBRR,pizzati2023physics} remove raindrops by leveraging the physics of raindrops or utilizing generated synthetic data.
However, the effectiveness of methods trained on synthetic data relies on the quality of raindrop rendering models.
Currently, raindrop rendering models commonly employ surface modeling and ray tracing~\cite{roser2009video,2010Realistic,hao2019learning}. Unfortunately, these methods may not fully consider factors such as scene depth, lighting conditions, and the 3D shapes of raindrops~\cite{roser2009video,pizzati2023physics}, which are crucial for realistic raindrop rendering.
Additionally, raindrops can introduce refraction effects~\cite{you2013adherent,halimeh2009raindrop}, distorting the shapes and positions of objects, which further complicates accurate raindrop rendering.
To the best of our knowledge, developing a rendering approach that can produce physically accurate raindrops adaptable to various scenes is challenging.

Motivated by the aforementioned challenges and the observation that existing raindrop datasets predominantly consist of background-focused images while neglecting raindrop-focused images and nighttime raindrops, we collect a novel dataset, called ``Raindrop Clarity".
Our dataset comprises 6,742 (1,836 daytime, 4,906 nighttime)  pairs and 8,444 (3,606 daytime, 4,838 nighttime) triplets of raindrop-focused images and background-focused images. 
All of them have the corresponding clear background images. 
The triplets include one more raindrop-free blurry background images.
Fig.~\ref{fig:data} shows samples and explanation on the meaning of our pairs and triplets.
From the 15,186 raindrops pairs/triplet, 5,442 are captured during daytime and the 9,744 at nighttime.
Raindrops Clarity exhibits diverse sizes and shapes, and raindrop appearance varies based on the scene and lighting conditions.
To our knowledge, Raindrop Clarity is the largest real-world raindrop dataset that contains both the pair and triplets, day and night.
Our main contributions are summarized as follows:
\begin{itemize}
	\item We present the Raindrop Clarity dataset, a large-scale real-world dataset that encompasses both raindrop-focused triplets and background-focused pairs.
	For the pairs, we provide images with blurry raindrops and their corresponding clear backgrounds.
	For the triplets, we additionally provide the raindrop-free blurry background. 
	\item Raindrop Clarity includes images collected under both daytime and nighttime.
	It aims to remove raindrops and recover clear backgrounds regardless of day or night, whether the raindrops or background are blurry.
	\item Our experimental results indicate that even with the application of state-of-the-art raindrop removal methods, there are instances where these methods fail. This suggests the existence of unresolved challenges in the field of raindrop removal, which await further attention and investigation.
\end{itemize}

\section{Related Work}
\label{sec:related}
\noindent\textbf{Raindrop Dataset.}
Qian~\etal~\cite{qian2018attentive} introduce a raindrop dataset consisting of raindrop and corresponding ground truths. 
Porav~\etal~\cite{porav2019can} propose a stereo RobotCar dataset, using a rain-making device that causes one lens to be affected by raindrops, while the other lens remains clear. 
Hao~\etal~\cite{hao2019learning} create a synthetic dataset of adherent raindrops with pixel-level masks for training deraining algorithms. 
Quan~\etal~\cite{quan2021removing} introduce RainDS, a dataset that includes raindrops, rain steaks, and their corresponding ground truths.
Soboleva and Shipitko~\cite{soboleva2021raindrops} introduce Windshields, a dataset containing raindrop images on both camera lenses and windshields, annotated with binary masks representing the raindrop areas. 
However, all of these datasets primarily consist of daytime images taken by cameras focusing on the background scenes and thus have blurry raindrops.

\noindent\textbf{Single Image Raindrop Removal.}
Most image deraining methods have primarily focused on removing rain streaks rather than raindrops. 
However, recent learning-based methods have been proposed specifically for single-image raindrop removal. 
Eigen~\etal~\cite{eigen2013restoring} introduce a supervised CNN-based method to remove raindrops, using a shallow CNN.
Qian~\etal~\cite{qian2018attentive} propose an AttentiveGAN for raindrop removal, that relies on attention maps but has a limited focus on local spatial information.
Quan~\etal~\cite{quan2019deep} propose RaindropAttention, which employs an attention mechanism to generate attention maps based on mathematical raindrop shapes, yet it lacks exploration of long-range information.
Liu~\etal~\cite{liu2019dual} demonstrate a dual residual network, by leveraging the complementary nature of paired residual blocks.
Hao~\etal~\cite{hao2019learning} propose a raindrop synthesis algorithm, used their synthetic dataset to train a multi-task network for raindrop detection and removal.
Porav~\etal~\cite{porav2019can} propose a Pix2PixHD-based deraining network architecture.
Subsequent work by Quan~\etal~\cite{quan2021removing} propose a cascaded network architecture for simultaneous removal of rain streaks and raindrops, achieved through a neural architecture search.
Yasarla and Patel~\cite{yasarla2019uncertainty} propose the uncertainty guided multi-scale residual learning network to effectively remove rain from images.
Zhang~\etal~\cite{zhang2021dual} propose the Dual Attention-in-Attention Model to remove rain streaks and raindrops. 

\noindent\textbf{Restoration Backbone.}
There are image restoration and enhancement tasks that utilize transformer-based models, taking advantage of the long-range dependencies, including IPT~\cite{chen2021pre} (vanilla transformer), Uformer~\cite{Uformer}, SwinIR~\cite{liang2021swinir} and Restormer~\cite{Restormer}. 
Xiao~\etal~\cite{xiao2022image} propose an image deraining transformer (IDT) for rain streak and raindrop removal with window-based and spatial-based
dual Transformer.
RaindropDiffusion~\cite{ozdenizci2023restoring} develops a UNet-based diffusion framework, which demonstrates robust generation capabilities for raindrops, rain streaks, and snow.
RainDiffusion~\cite{wei2023raindiffusion}, an unsupervised diffusion model, utilizes unpaired real-world data instead of weakly adversarial training. 
DiT~\cite{peebles2022scalable} is a transformer-based diffusion model.

\begin{figure}[t!]
	\centering
	\captionsetup[subfig]{labelformat=empty}
	{\includegraphics[width=0.7\linewidth]{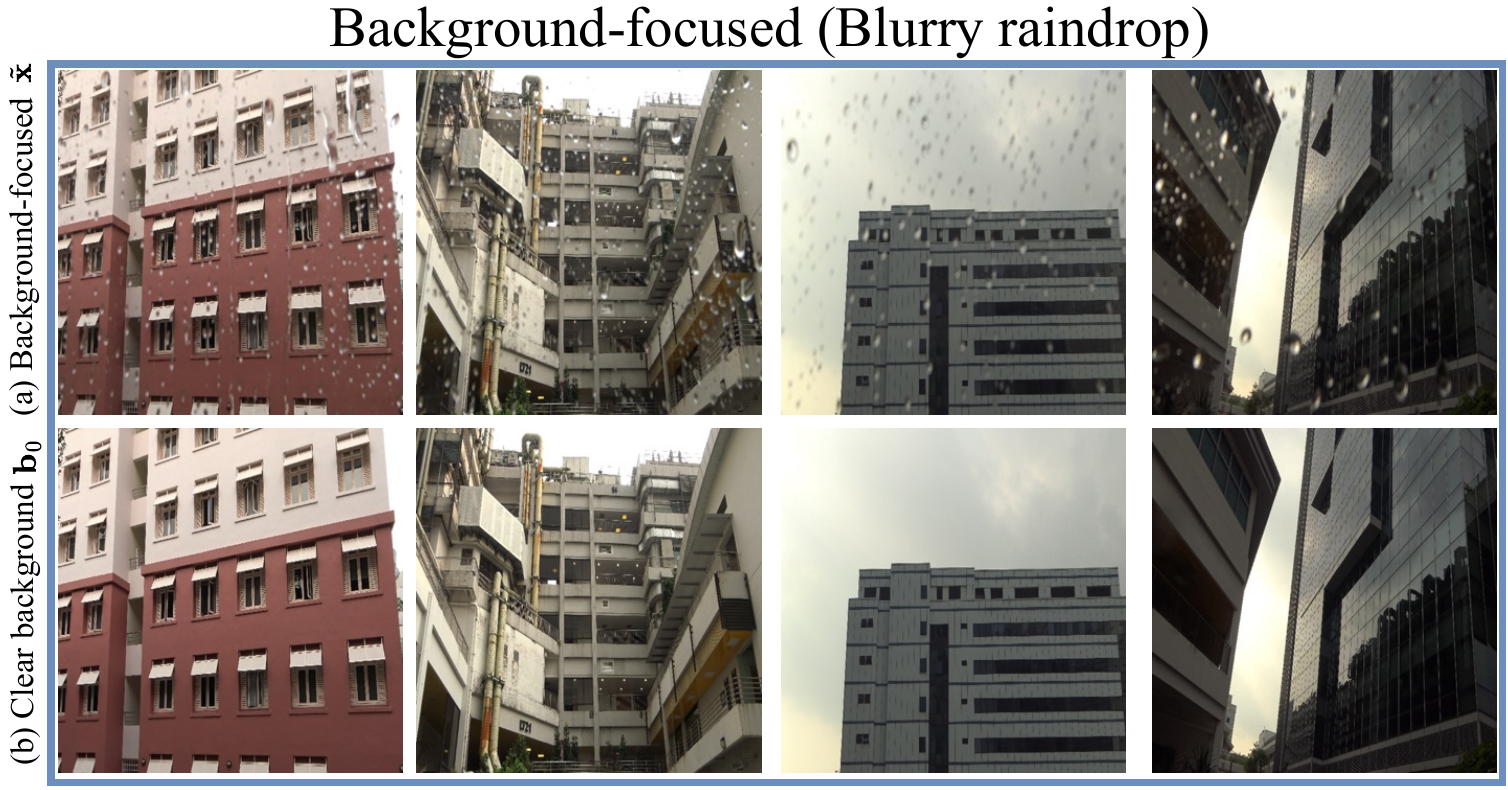}}\hfill
	\caption{Pairs $(\xw,\bb_0)$ show various sizes, shapes, and densities.}
	\label{fig:2variety}
\end{figure}

\section{Raindrop Clarity Dataset}
\begin{figure}[t!]
	\centering
	\captionsetup[subfig]{labelformat=empty}
	{\includegraphics[width=0.7\linewidth]{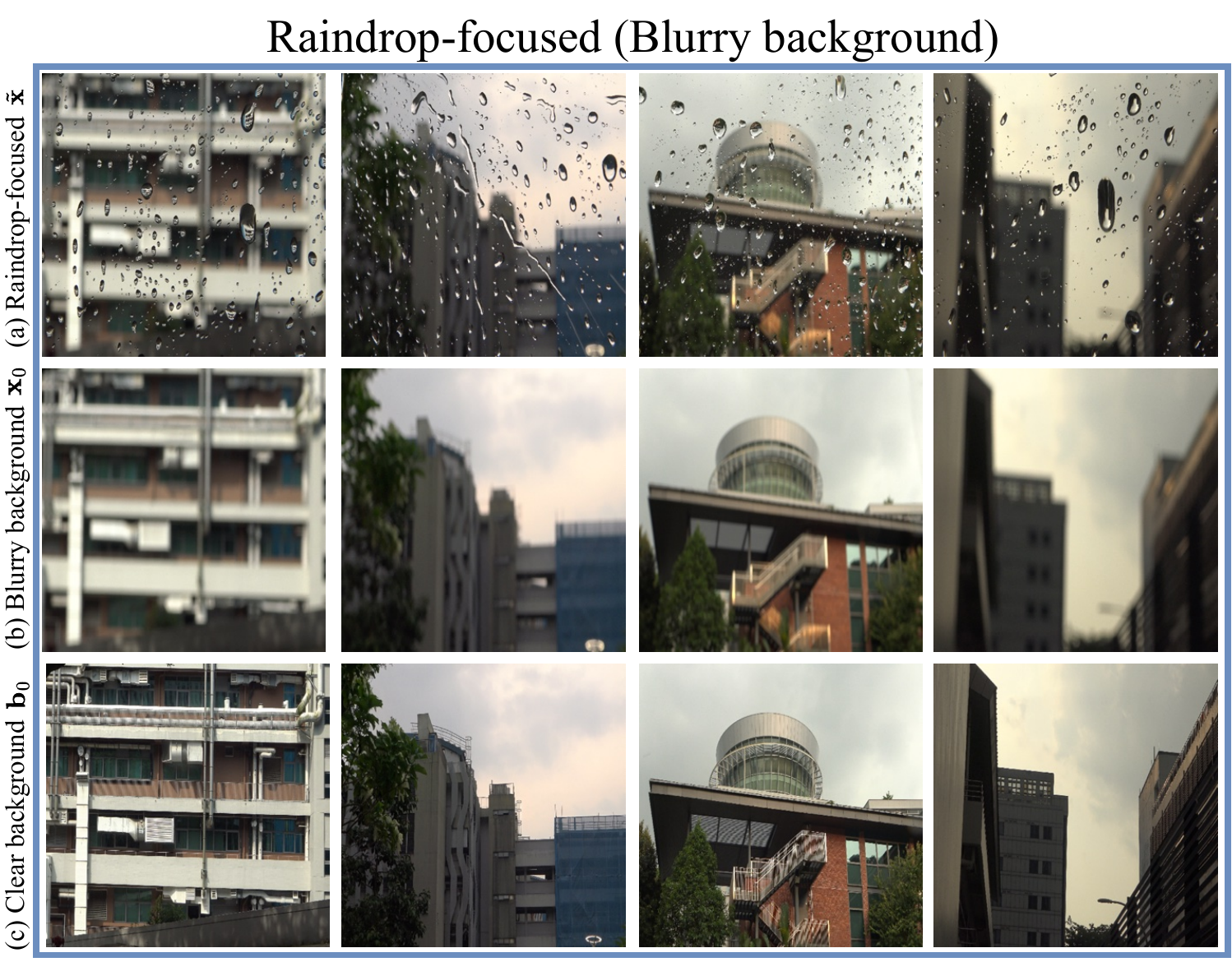}}\hfill
	\caption{The triplets $(\xw,\x_0,\bb_0)$ cover diverse scenes and shapes.}
	\label{fig:3variety}
\end{figure}

\subsection{Data Collection}
To create Raindrop Clarity, we employed a sphere pan-tilt platform to keep our camera stationary.
The images were captured using Sony FDR-AX33 4K Ultra HD Handycam Camcorder, Sony alpha 7R III digital camera, the back camera of an iPhone 14 Pro and iPhone 15 Pro Max, ensuring the highest quality possible. The 4K Camcorder captured RGB images at a resolution 3840$\times$2164.

Using the optical refraction model (raindrops appear upside-down), we projected raindrops by spraying water on glass and adjusting the distance, angle, speed, and patterns of the nozzle/wipers. We also captured in rain conditions.

Our collection procedure was designed to ensure diversity in the raindrop (such as focuses, shapes and sizes), background scenes and lighting conditions.
First, we fixed the camera position on a tripod and installed a glass plate. The distance between the camera and the glass plate ranged from 5cm to 25cm.
Second, we sprayed waterdrops/raindrops onto the glass plate and focused the camera on the raindrops. 
Third, we sprayed the waterdrops/raindrops to capture images of them under different shapes and sizes, ranging from sparse to dense. 
The captured images include elliptical raindrops and raindrop flow traces ($\xw$).
Fourth, we removed the glass plate, allowing the camera to focus on the very close plane. The duration is kept very short, within milliseconds, to minimize lighting changes. In this way, we obtained  raindrop-free blurry background images ($\x_0$). 
Finally, we adjusted the camera focus to the background, directing the camera to focus on the distant background ($\bb_0$).

\subsection{Characteristics of Raindrop Clarity}
Raindrop Clarity is a new dataset consisting of around 15,186 high-quality image raindrop pairs and triplets. 
In Fig.~\ref{fig:data}, our dataset introduces a unique element of triplet data, which includes raindrop-focused images, corresponding blurry raindrop-free background images, and their clear background equivalents, distinguishing it from existing datasets.
The second feature that distinguishes the Raindrop Clarity from the existing ones is the collection of night raindrops.

{\flushleft \bf Objectives.}
Our dataset is designed to support research in the elimination of raindrops and the restoration of a clear background, irrespective of whether the camera's focus is on the raindrops (leading to a blurry background) or on the background (resulting in blurry raindrops). This holds true regardless of whether it is daytime or nighttime.

{\flushleft \bf Raindrops and Diversity.}
Our Raindrop Clarity dataset offers an extensive coverage of diverse raindrop sizes, shapes, and occlusion types in various scenes and lighting conditions, making it an exceptionally comprehensive resource for advancing raindrop removal research.
In Fig.~\ref{fig:2variety} and Fig.~\ref{fig:3variety}, the dataset captures raindrops and rain flows of different sizes and shapes, including heavy raindrops that result in unpredictable water flows and blurry backgrounds when they hit window panes or car windshields.
Two primary types of occlusions caused by raindrops, namely partial and complete occlusions, are included in our dataset. Both types present unique challenges for raindrop removal algorithms.

{\flushleft \bf Scenes and Large-scale.}
The dataset encompasses a wide range of background scenes, such as urban, countryside, campus, wilderness, road, and aerial views, covering both day and night. This variation allows for the exploration of different lighting and weather conditions in different scenes. 
With approximately 38,816 images in total, including 15,186 high-quality pairs/triplets of images, the Raindrop Clarity dataset is one of the largest datasets of its kind.

\begin{figure}[t!]
	\centering
	\captionsetup[subfig]{labelformat=empty}
	\setcounter{subfigure}{0}
	{\includegraphics[width=0.7\linewidth]{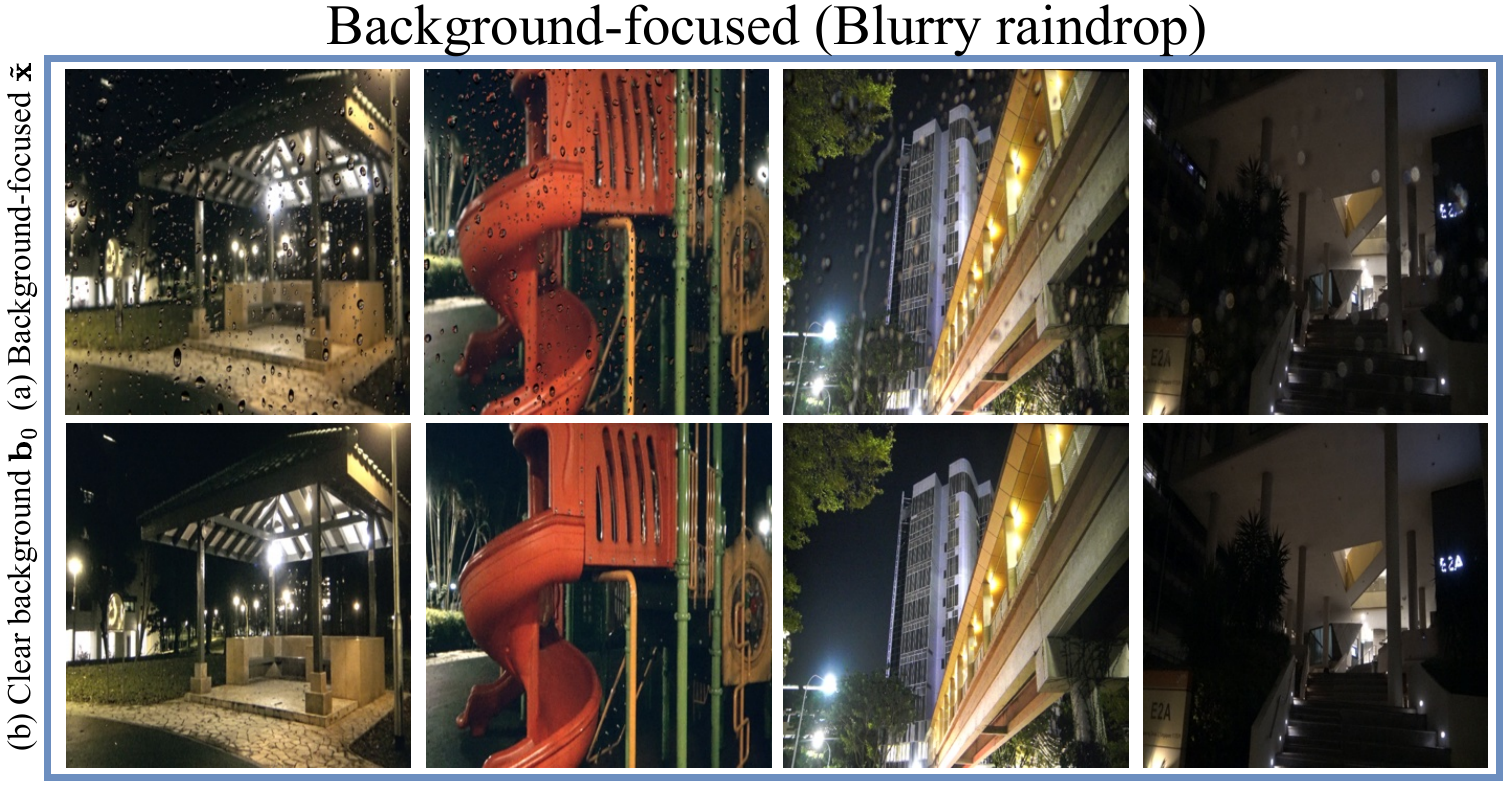}}\hfill
	\caption{The pairs $(\xw,\bb_0)$ under different lighting conditions.}
	\label{fig:2variety_n}
\end{figure}

\begin{figure}[t!]
	\centering
	\captionsetup[subfig]{labelformat=empty}
	\setcounter{subfigure}{0}
	{\includegraphics[width=0.7\linewidth]{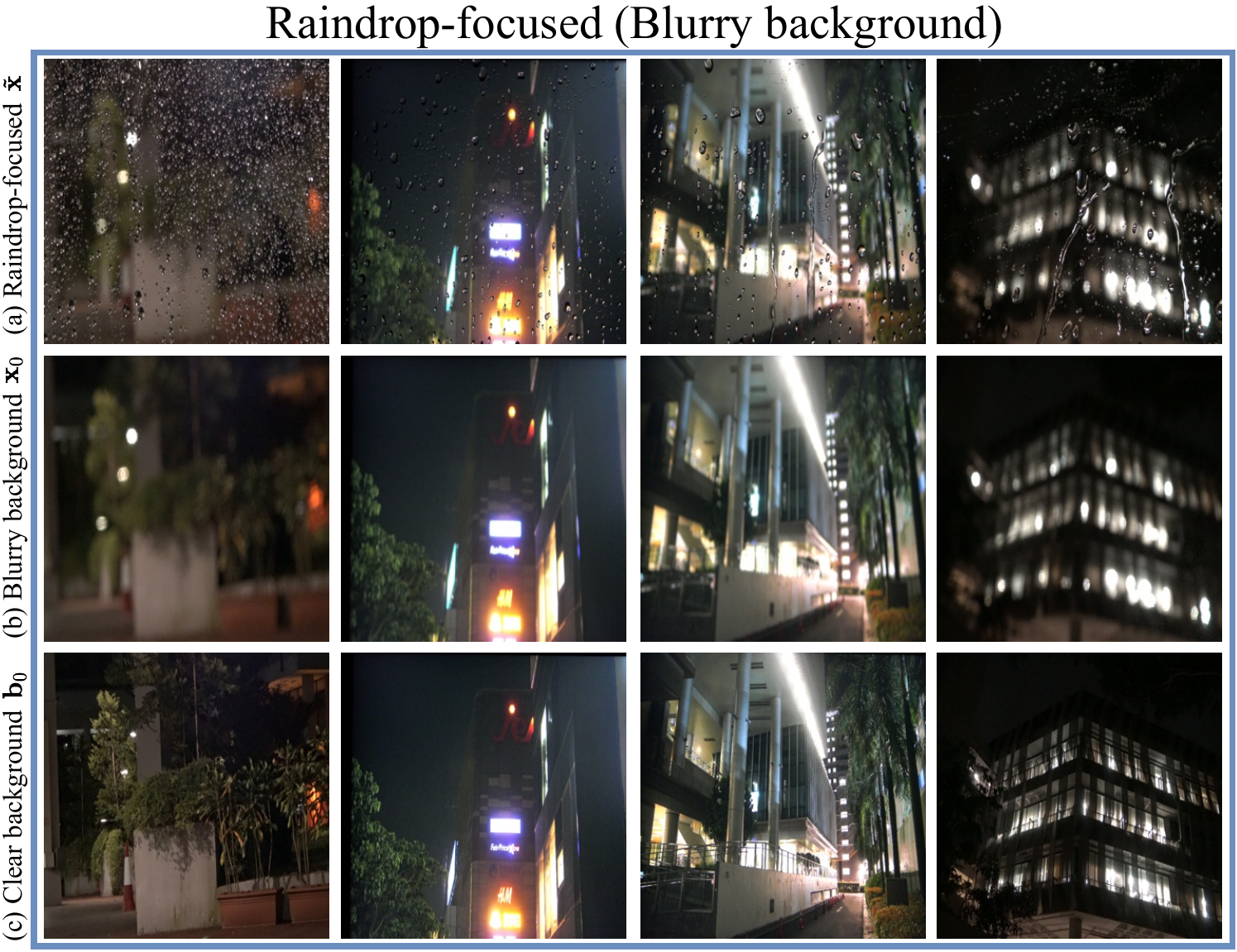}}\hfill
	\caption{Triplets $(\xw,\x_0,\bb_0)$ under different lighting conditions.}
	\label{fig:3variety_n}
\end{figure}

{\flushleft \bf Lighting Conditions.}
The appearance of raindrops highly depends on the surrounding environment.
A distinctive feature of our dataset is its comprehensive coverage of various environmental conditions, setting it apart from other datasets.
The dataset's variation in day and night conditions makes it valuable for developing raindrop removal algorithms.

During the daytime, raindrop images exhibit a diverse array of appearances influenced by multiple factors such as sunlight direction, intensity, and shadowing. 
In Figs.~\ref{fig:2variety}-\ref{fig:3variety}, our dataset captures this variety, with images taken at different times of the day and under varying weather conditions (such as light rain, heavy rain, and rain combined with wind), providing rich information for algorithms to learn.

Unlike daytime, nighttime raindrops present unique challenges due to artificial lighting conditions and low light.
The appearance of raindrops can drastically change under street lights, headlights, or neon lights. 
In Fig.~\ref{fig:2variety_n} and Fig.~\ref{fig:3variety_n}, our dataset includes night images with varied artificial lights, contributing to the complexity and comprehensiveness of the dataset. These night images are challenging due to the higher contrast between raindrops and the background, and potential overexposure or underexposure issues.

\begin{table*}[t!]
	\caption{Summary of comparisons between our Raindrop Clarity and existing datasets. Bg-focus = Background-focused, Rd-focus = Raindrop-focused, RD = Raindrop. Raindrop Clarity uniquely features raindrop-focused triplets and the nighttime raindrops.}
	\centering
	\resizebox{\textwidth}{!}{
		\begin{tabular}{c|c|c|c|c|c|c|c}\hline
			\multirow{2}{*}{Datasets}                          &\multirow{2}{*}{Components}   &\multirow{2}{*}{N. Images}      &Syn/ &\multicolumn{2}{c|}{Bg-focus} &\multicolumn{2}{c}{Rd-focus}\\\cline{5-8}
			&  &   &Real    &Day  &Night &Day &Night\\\hline
			Our RD Clarity (Day)   &3,606 Triplets+1,836 Pairs &14,490       &Real      &\checkmark  &$\times$ 
			&\checkmark &$\times$\\
			Our RD Clarity (Night) &4,838 Triplets+4,906 Pairs  &24,326     &Real      &$\times$  &\checkmark 
			&$\times$ &\checkmark\\\hline
			Raindrop Qian~\cite{qian2018attentive}   &919 Pairs  &1,838       &Real	  &\checkmark  &$\times$ &$\times$ &$\times$\\\hline
			RainDS-Real~\cite{quan2021removing}	&248 Quadruplets &992  &Real      &\checkmark  &$\times$ &$\times$ &$\times$\\\hline
			RainDS-Syn~\cite{quan2021removing}	&1,200 Quadruplets &4,800 &Syn      &\checkmark  &$\times$ &$\times$ &$\times$\\\hline
			RobotCar Porav~\cite{porav2019can}		&4,818 Pairs   &9,636         &Real     &\checkmark  &$\times$ &$\times$ &$\times$\\\hline 
			Windshield~\cite{soboleva2021raindrops} &3,390     &3,390     &Real   &\checkmark  &$\times$ &$\times$ &$\times$\\\hline
	\end{tabular}}
	\label{tb:compare}
\end{table*}

\begin{figure}[t!]
	\captionsetup[subfig]{labelformat=empty}
	\centering
		{\includegraphics[width=0.7\linewidth]{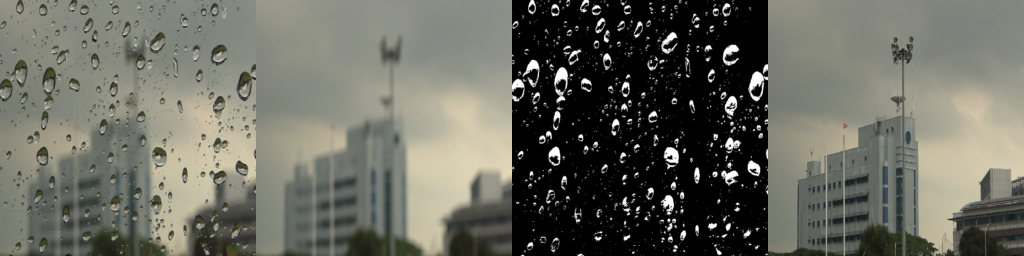}}\hfill
		{\includegraphics[width=0.7\linewidth]{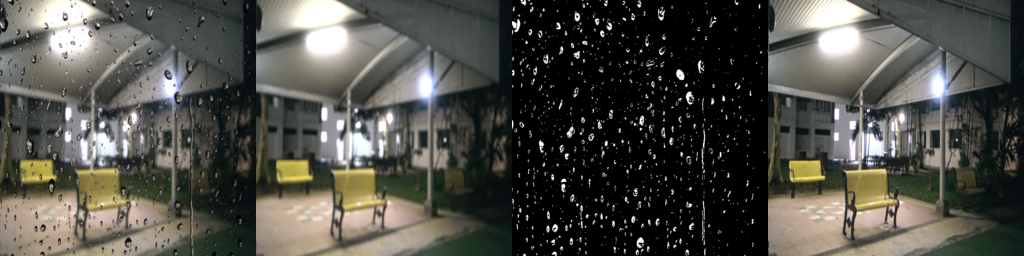}}\hfill
	\caption{We show $(\xw,\x_0,\mathbf{\tilde{m}},\bb_0)$, difference map $\mathbf{\tilde{m}}$ is derived by distinguishing raindrop-focused $\xw$ from blurry background $\x_0$.}
	\label{fig:mask}
\end{figure}

\begin{figure*}[t!]
	\centering
		\captionsetup[subfigure]{labelformat=empty}
		\setcounter{subfigure}{0}
		\subfloat{\includegraphics[width=0.164\textwidth]{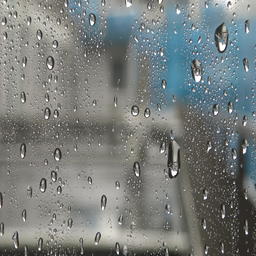}}\hfill	
		\subfloat{\includegraphics[width=0.164\textwidth]{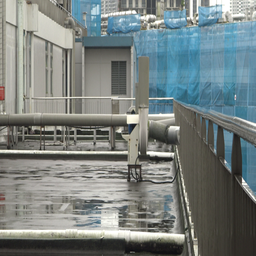}}\hfill
		\subfloat{\includegraphics[width=0.164\textwidth]{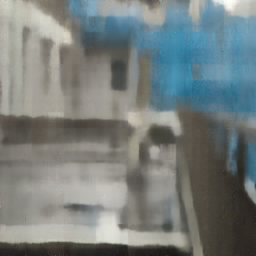}}\hfill
		\subfloat{\includegraphics[width=0.164\textwidth]{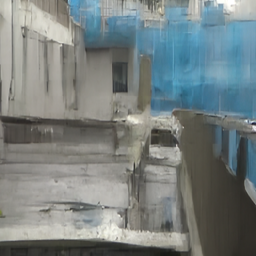}}\hfill
		\subfloat{\includegraphics[width=0.164\textwidth]{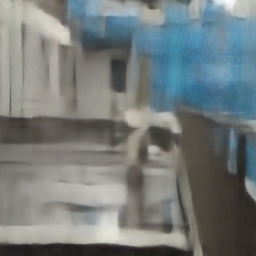}}\hfill
		\subfloat{\includegraphics[width=0.164\textwidth]{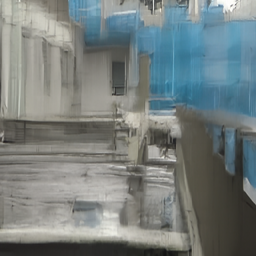}}\hfill
		\setcounter{subfigure}{0}
		\subfloat{\includegraphics[width=0.164\textwidth]{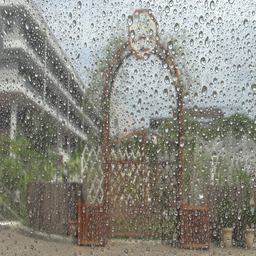}}\hfill	
		\subfloat{\includegraphics[width=0.164\textwidth]{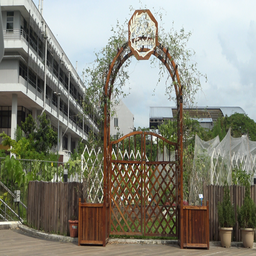}}\hfill
		\subfloat{\includegraphics[width=0.164\textwidth]{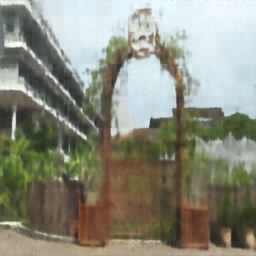}}\hfill
		\subfloat{\includegraphics[width=0.164\textwidth]{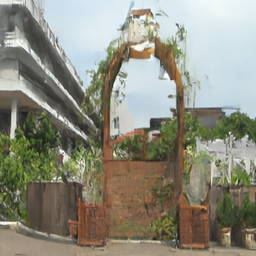}}\hfill
		\subfloat{\includegraphics[width=0.164\textwidth]{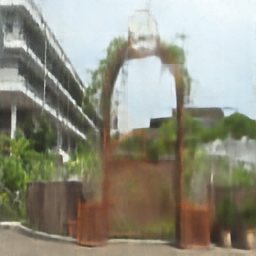}}\hfill
		\subfloat{\includegraphics[width=0.164\textwidth]{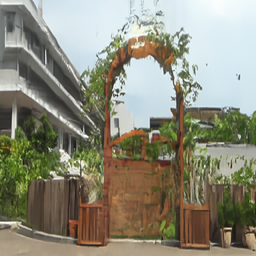}}\hfill
		\setcounter{subfigure}{0}
		\subfloat{\includegraphics[width=0.164\textwidth]{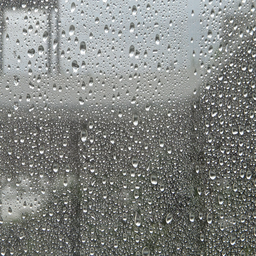}}\hfill	
		\subfloat{\includegraphics[width=0.164\textwidth]{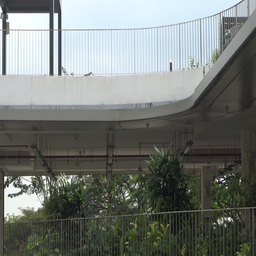}}\hfill
		\subfloat{\includegraphics[width=0.164\textwidth]{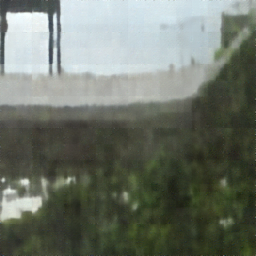}}\hfill
		\subfloat{\includegraphics[width=0.164\textwidth]{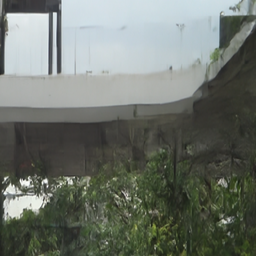}}\hfill
		\subfloat{\includegraphics[width=0.164\textwidth]{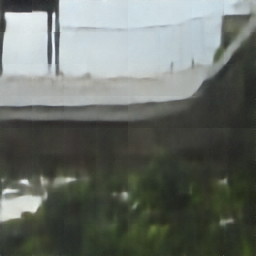}}\hfill
		\subfloat{\includegraphics[width=0.164\textwidth]{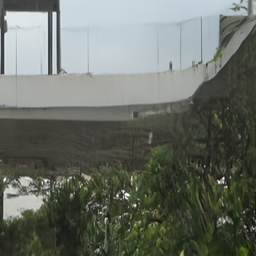}}\hfill
		\setcounter{subfigure}{0}
		\subfloat[Input]{\includegraphics[width=0.164\textwidth]{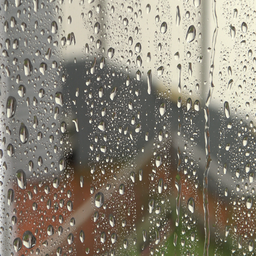}}\hfill	
		\subfloat[Ground Truth]{\includegraphics[width=0.164\textwidth]{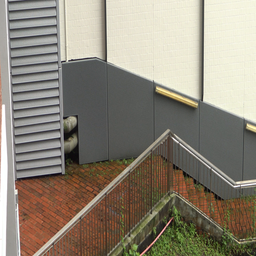}}\hfill
		\subfloat[IDT~\cite{xiao2022image}]{\includegraphics[width=0.164\textwidth]{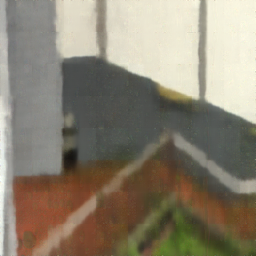}}\hfill
		\subfloat[RDiffusion~\cite{ozdenizci2023restoring}]{\includegraphics[width=0.164\textwidth]{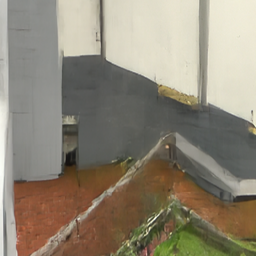}}\hfill
		\subfloat[Restormer~\cite{Restormer}]{\includegraphics[width=0.164\textwidth]{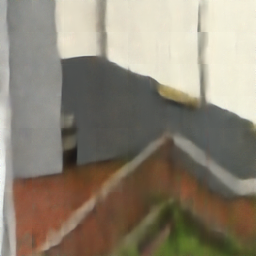}}\hfill
		\subfloat[DiT~\cite{peebles2022scalable}]{\includegraphics[width=0.164\textwidth]{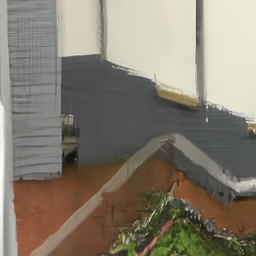}}\hfill
	\caption{Comparison of existing methods on daytime raindrop-focused images, covering both raindrop algorithms and restoration backbones.}
	\label{fig:dayrd}
\end{figure*} 

{\flushleft \bf Background- and Raindrop-focused.}
When the raindrop image $\xw$ is background-focused, we decompose a raindrop degraded image $\xw$ into a clear background image $\bb_0$ by following this model~\cite{qian2018attentive}: 
$\xw = (1 - \mathbf{M}) \odot \x_0 + \mathbf{D}\label{eq:model1}$, where $\mathbf{D}$ is the effect of the blurry raindrops, referring to the localized blurry imagery formed by the light reflected from the surroundings. The operator $\odot$ denotes element-wise multiplication. 
When binary mask $\mathbf{M(x)}$ equals 1, it signifies that the pixel $x$ belongs to the area affected by raindrops. Conversely, if $\mathbf{M(x)}$ does not equal 1, it indicates that the pixel is within the background areas.
In the case where $\xw$ is background-focused raindrop images, $\x_0$ is identical to the clear, raindrop-free image $\bb_0$, namely $\x_0 = \bb_0$.
Therefore, we use $(\xw,\bb_0)$ to stand for background-focused (resulting in blurry raindrops) pairs, shown in the left part of Fig.~\ref{fig:data}.

In the right part of Fig.~\ref{fig:data} and Fig.~\ref{fig:mask}, when $\xw$ is the raindrop-focused image, $\x_0$ is the corresponding raindrop-free blurry background image, while $\bb_0$ is the clear background image. In this case, $\x_0 \neq \bb_0$.
We can calculate the pixel-level difference map $\mathbf{\tilde{m}}$ between the raindrop-free blurry background $\x_0$ and the raindrop-focused $\xw$ as:
$\mathbf{\tilde{m}} = \xw - \x_0 \label{eq:model2}$.
Therefore, we use $(\xw,\x_0,\bb_0)$ to stand for raindrop-focused (resulting in blurry background) triplets.

\begin{table*}[!t]
	\caption{Quantitative evaluation of daytime and nighttime raindrop data with a benchmark for raindrop algorithms and restoration backbones.}
	\label{tab:PSNR_SSIM} 
	\centering
	\resizebox{\textwidth}{!}{
		\begin{tabular}{@{}llcccccc|cccc@{}}\toprule
			\multicolumn{2}{c}{\multirow{2}{*}{Data\textbackslash{}Method}}  &\multicolumn{1}{c}{\multirow{2}{*}{Input}} &\multicolumn{4}{c}{Raindrop Algorithms} &  &\multicolumn{3}{c}{{Restoration Backbones}} \\ \cmidrule(l){4-11} 
			\multicolumn{2}{c}{}                                              &\multicolumn{1}{c}{}  &AtGAN~\cite{qian2018attentive} &Robot~\cite{porav2019can} &RainDS~\cite{quan2021removing} &IDT~\cite{xiao2022image} &RDdiff~\cite{ozdenizci2023restoring}  &Restor.~\cite{Restormer} &Uform.~\cite{Uformer} &DiT~\cite{peebles2022scalable}        \\ \midrule
			\multirow{3}{*}{Day}&PSNR$\uparrow$   &21.92        &23.62     &23.82   &25.39   &26.05  &25.52          &\bf{26.08}                   &25.78     &26.03 \\ \cmidrule(l){2-11} 
			&SSIM$\uparrow$                                       &0.560        &0.658    &0.670   &0.717   &0.736  &0.734         & 0.748                     &0.727     &\textbf{0.752} \\ \cmidrule(l){2-11} 
			&LPIPS$\downarrow$   							   &0.247        &0.200      &0.191   &0.147    &0.141  &0.111         & 0.131                   &0.148      & {\textbf{0.106}} \\ \midrule
			\multirow{3}{*}{Night}&PSNR$\uparrow$ &24.78        &24.38      &24.57  &25.39   &26.81  &26.48          &\bf{26.87}  &25.26      &26.23\\ \cmidrule(l){2-11} 
			&SSIM$\uparrow$                                        & 0.726        &0.773   &0.787   &0.820  &\textbf{0.851}  &0.831        &0.848                    &0.816     &0.826\\ \cmidrule(l){2-11} 
			&LPIPS$\downarrow$                                  & 0.209        &0.185  &0.182    & 0.139  &0.125   &0.112         &0.123                      & 0.142      &\textbf{0.111} \\ \bottomrule
	\end{tabular}}
\end{table*}

\begin{figure*}[t!]
	\centering
		\captionsetup[subfigure]{labelformat=empty}
		\setcounter{subfigure}{0}
		\subfloat{\includegraphics[width=0.164\textwidth]{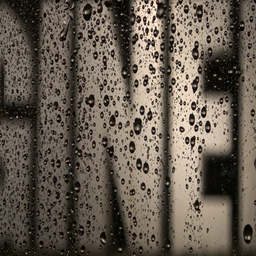}}\hfill	
		\subfloat{\includegraphics[width=0.164\textwidth]{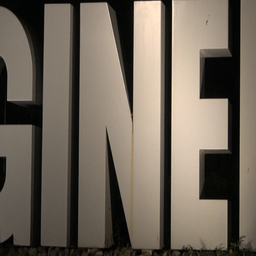}}\hfill
		\subfloat{\includegraphics[width=0.164\textwidth]{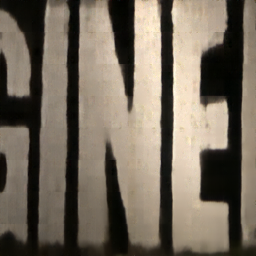}}\hfill
		\subfloat{\includegraphics[width=0.164\textwidth]{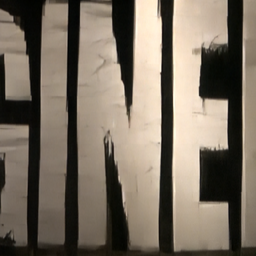}}\hfill
		\subfloat{\includegraphics[width=0.164\textwidth]{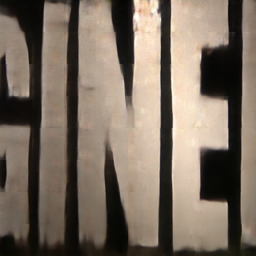}}\hfill
		\subfloat{\includegraphics[width=0.164\textwidth]{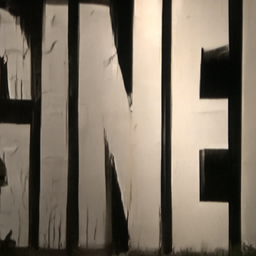}}\hfill
		\setcounter{subfigure}{0}
		\subfloat{\includegraphics[width=0.164\textwidth]{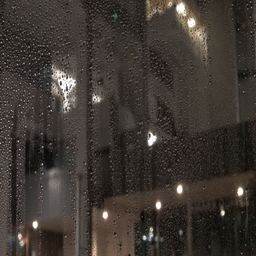}}\hfill	
		\subfloat{\includegraphics[width=0.164\textwidth]{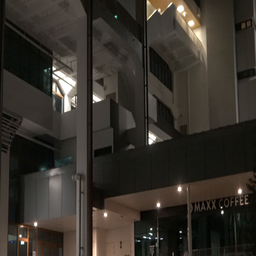}}\hfill
		\subfloat{\includegraphics[width=0.164\textwidth]{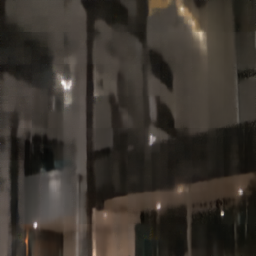}}\hfill
		\subfloat{\includegraphics[width=0.164\textwidth]{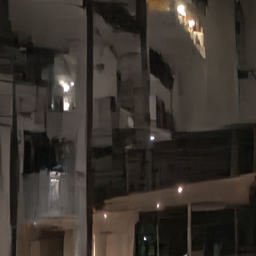}}\hfill
		\subfloat{\includegraphics[width=0.164\textwidth]{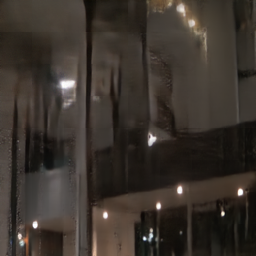}}\hfill
		\subfloat{\includegraphics[width=0.164\textwidth]{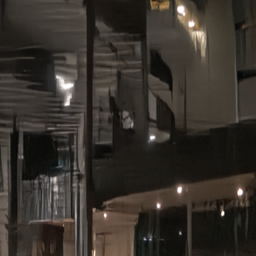}}\hfill
		\setcounter{subfigure}{0}
		\subfloat{\includegraphics[width=0.164\textwidth]{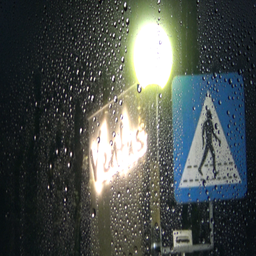}}\hfill	
		\subfloat{\includegraphics[width=0.164\textwidth]{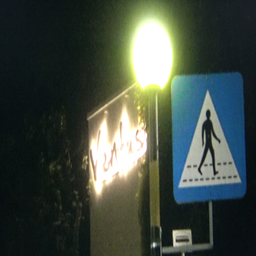}}\hfill
		\subfloat{\includegraphics[width=0.164\textwidth]{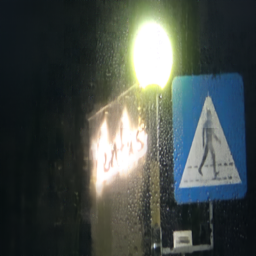}}\hfill
		\subfloat{\includegraphics[width=0.164\textwidth]{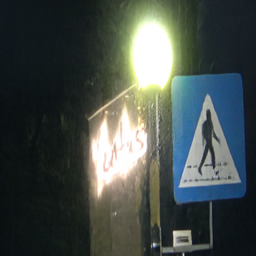}}\hfill
		\subfloat{\includegraphics[width=0.164\textwidth]{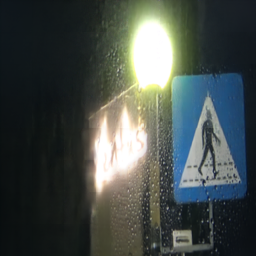}}\hfill
		\subfloat{\includegraphics[width=0.164\textwidth]{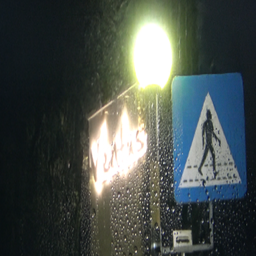}}\hfill
		\setcounter{subfigure}{0}
		\subfloat[Input]{\includegraphics[width=0.164\textwidth]{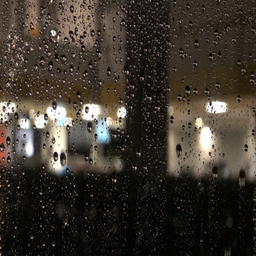}}\hfill
		\subfloat[Ground Truth]{\includegraphics[width=0.164\textwidth]{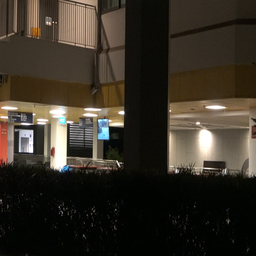}}\hfill
		\subfloat[IDT~\cite{xiao2022image}]{\includegraphics[width=0.164\textwidth]{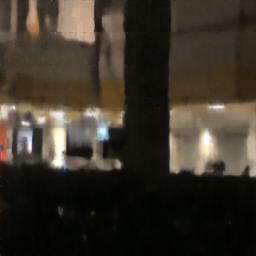}}\hfill
		\subfloat[RDiffusion~\cite{ozdenizci2023restoring}]{\includegraphics[width=0.164\textwidth]{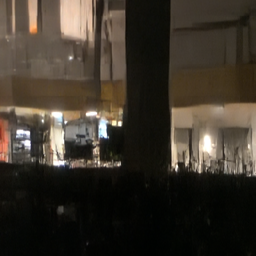}}\hfill
		\subfloat[Restormer~\cite{Restormer}]{\includegraphics[width=0.164\textwidth]{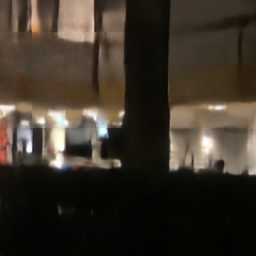}}\hfill
		\subfloat[DiT~\cite{peebles2022scalable}]{\includegraphics[width=0.164\textwidth]{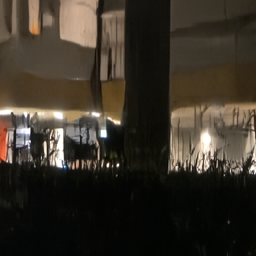}}\hfill
	\caption{Comparison of methods on nighttime raindrop-focused images, covering both raindrop algorithms and restoration backbones.}
	\label{fig:nightrd}
\end{figure*} 

Since the camera is stationary, neither the pairs $(\xw,\bb_0)$ nor the triplets $(\xw,\x_0,\bb_0)$ have alignment problems. This is also evident in the raindrop difference maps $\mathbf{\tilde{m}}$, which accurately assign values to raindrop regions across multiple images.
Specifically, for each pixel $\mathbf{x}$ in the input image, any discrepancies in the map's assigned values for the static background would indicate alignment issues; however, these are not observed, as $\mathbf{\tilde{m}}$ indicates only the raindrop regions.
Since the pairs/triplets of images are captured in quick succession, there are no problems with changes in illumination.

The captured images do not contain any personal information such as phone numbers, or human faces.
We will release the dataset publicly with car license plates obscured, as we believe that our Raindrop Clarity dataset will benefit the progress of the raindrop removal field.

\begin{figure*}[t!]
	\centering
		\captionsetup[subfigure]{labelformat=empty}
		\setcounter{subfigure}{0}
		\subfloat{\includegraphics[width=0.164\textwidth]{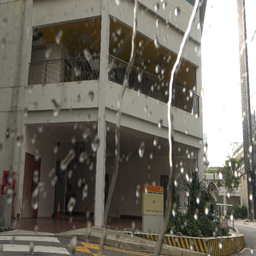}}\hfill	
		\subfloat{\includegraphics[width=0.164\textwidth]{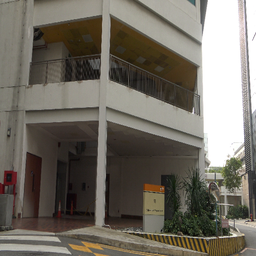}}\hfill
		\subfloat{\includegraphics[width=0.164\textwidth]{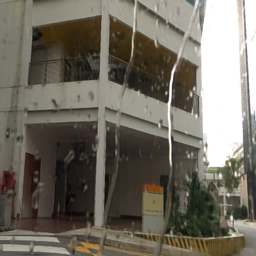}}\hfill
		\subfloat{\includegraphics[width=0.164\textwidth]{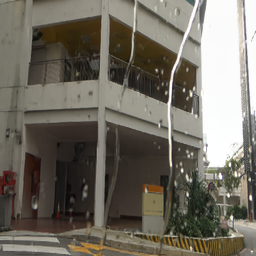}}\hfill
		\subfloat{\includegraphics[width=0.164\textwidth]{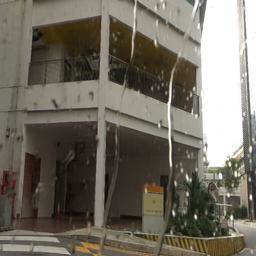}}\hfill
		\subfloat{\includegraphics[width=0.164\textwidth]{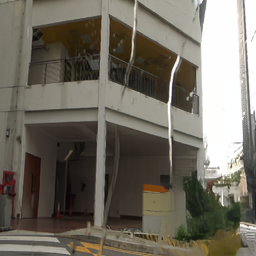}}\hfill
		\setcounter{subfigure}{0}
		\subfloat{\includegraphics[width=0.164\textwidth]{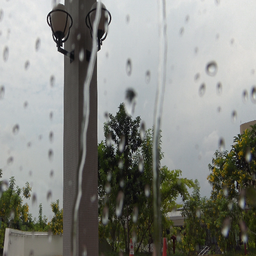}}\hfill	
		\subfloat{\includegraphics[width=0.164\textwidth]{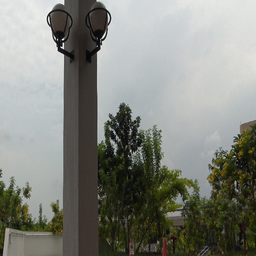}}\hfill
		\subfloat{\includegraphics[width=0.164\textwidth]{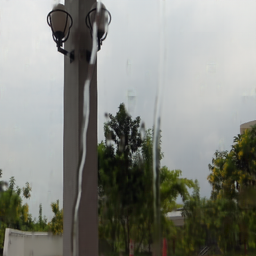}}\hfill
		\subfloat{\includegraphics[width=0.164\textwidth]{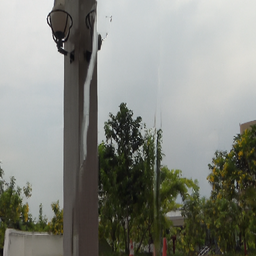}}\hfill
		\subfloat{\includegraphics[width=0.164\textwidth]{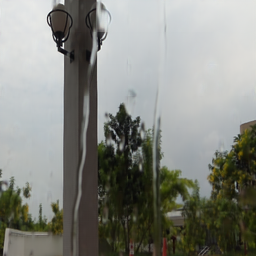}}\hfill
		\subfloat{\includegraphics[width=0.164\textwidth]{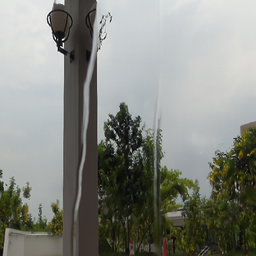}}\hfill
		\setcounter{subfigure}{0}
		\subfloat[Input]{\includegraphics[width=0.164\textwidth]{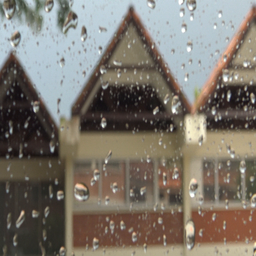}}\hfill	
		\subfloat[Ground Truth]{\includegraphics[width=0.164\textwidth]{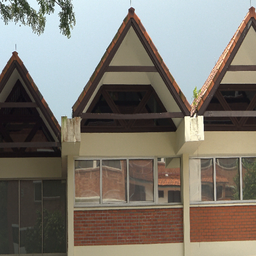}}\hfill
		\subfloat[IDT~\cite{xiao2022image}]{\includegraphics[width=0.164\textwidth]{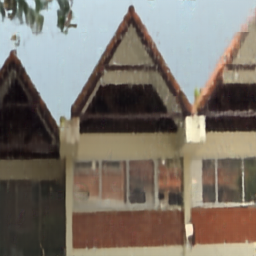}}\hfill
		\subfloat[RDiffusion~\cite{ozdenizci2023restoring}]{\includegraphics[width=0.164\textwidth]{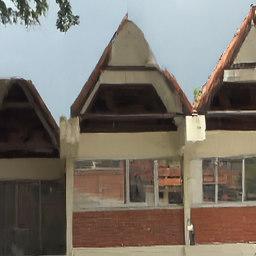}}\hfill
		\subfloat[Restormer~\cite{Restormer}]{\includegraphics[width=0.164\textwidth]{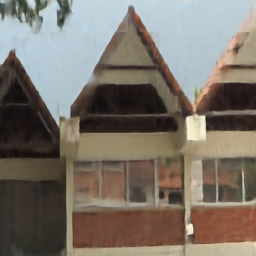}}\hfill
		\subfloat[DiT~\cite{peebles2022scalable}]{\includegraphics[width=0.164\textwidth]{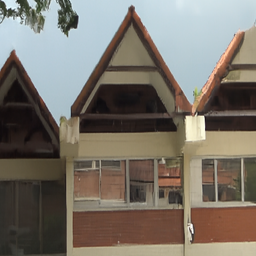}}\hfill
	\caption{Comparison of methods on daytime background-focused images, covering both raindrop algorithms and restoration backbones.}
	\label{fig:daybd}
\end{figure*} 

\section{Experiments}
\label{sec:experiments}

{\flushleft \bf Experimental Setup.}
Our benchmark includes two parts: Daytime and Nighttime Raindrop Removal.
As shown in Table \ref{tb:compare}, our daytime raindrop dataset contains a total of 5,442 paired/triplets images, where 4,713 paired images are used for the training set, and the remaining 729 paired images are regarded as the test set.
Our nighttime raindrop dataset consists of 9,744 paired/triplets images, where 8,655 paired images are selected as the training set and the rest 1,089 paired images are used for evaluation.

{\flushleft \bf  Baselines.}
To create the benchmark, we retrain existing state-of-the-art methods on our daytime and nighttime raindrop datasets. 
Our benchmark consists of five raindrop algorithms~\cite{qian2018attentive, porav2019can, quan2021removing, xiao2022image, ozdenizci2023restoring} and three restoration methods~\cite{Restormer, Uformer, peebles2022scalable}. 
Specifically, IDT~\cite{xiao2022image}, RDiffusion~\cite{ozdenizci2023restoring}, RainDS~\cite{quan2021removing}, Robot Car~\cite{porav2019can} and AtGAN~\cite{qian2018attentive} are raindrop algorithms, while Restormer~\cite{Restormer}, Uformer~\cite{Uformer} and DiT~\cite{peebles2022scalable} are restoration backbones.
For a fair comparison, we conduct experiments using the original implementation of these methods. All experiments are implemented in PyTorch and are trained on four NVIDIA RTX A5000 GPUs.

\begin{figure*}[t!]
	\centering
		\captionsetup[subfigure]{labelformat=empty}
		\setcounter{subfigure}{0}
		\subfloat{\includegraphics[width=0.164\textwidth]{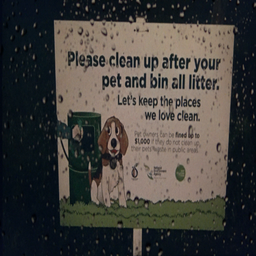}}\hfill	
		\subfloat{\includegraphics[width=0.164\textwidth]{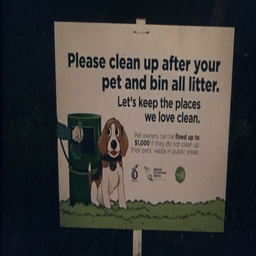}}\hfill
		\subfloat{\includegraphics[width=0.164\textwidth]{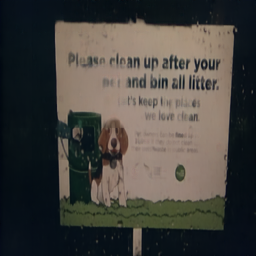}}\hfill
		\subfloat{\includegraphics[width=0.164\textwidth]{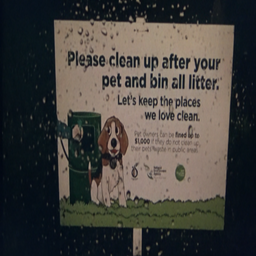}}\hfill
		\subfloat{\includegraphics[width=0.164\textwidth]{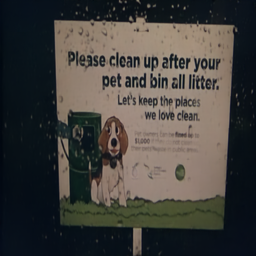}}\hfill
		\subfloat{\includegraphics[width=0.164\textwidth]{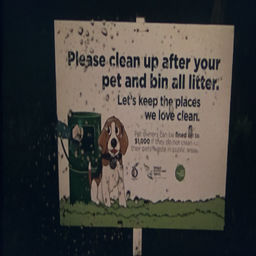}}\hfill
		\setcounter{subfigure}{0}
		\subfloat{\includegraphics[width=0.164\textwidth]{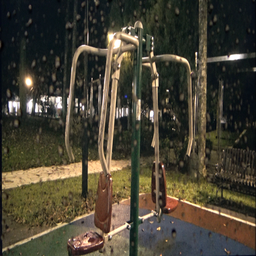}}\hfill	
		\subfloat{\includegraphics[width=0.164\textwidth]{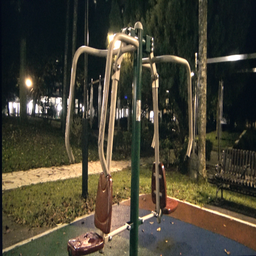}}\hfill
		\subfloat{\includegraphics[width=0.164\textwidth]{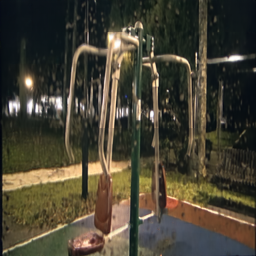}}\hfill
		\subfloat{\includegraphics[width=0.164\textwidth]{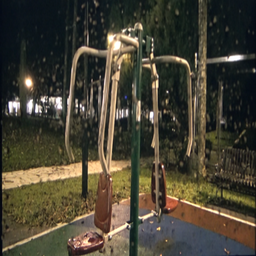}}\hfill
		\subfloat{\includegraphics[width=0.164\textwidth]{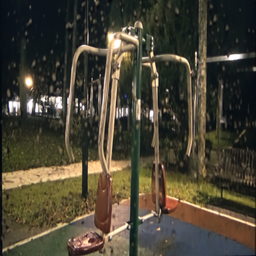}}\hfill
		\subfloat{\includegraphics[width=0.164\textwidth]{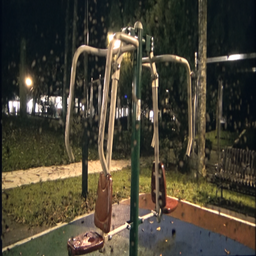}}\hfill
			\setcounter{subfigure}{0}
			\subfloat{\includegraphics[width=0.164\textwidth]{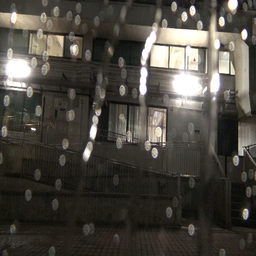}}\hfill	
			\subfloat{\includegraphics[width=0.164\textwidth]{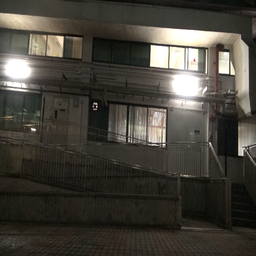}}\hfill
			\subfloat{\includegraphics[width=0.164\textwidth]{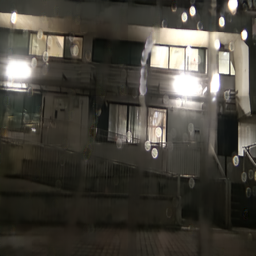}}\hfill
			\subfloat{\includegraphics[width=0.164\textwidth]{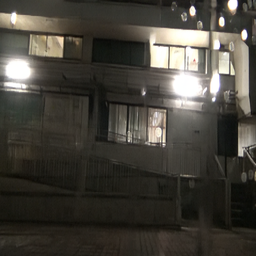}}\hfill
			\subfloat{\includegraphics[width=0.164\textwidth]{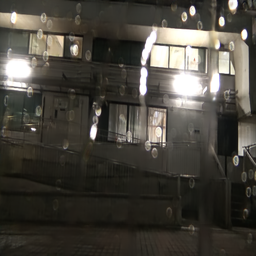}}\hfill
			\subfloat{\includegraphics[width=0.164\textwidth]{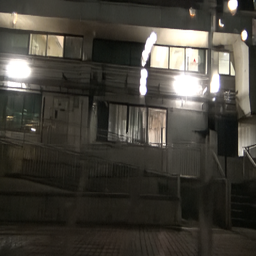}}\hfill
		\subfloat{\includegraphics[width=0.164\textwidth]{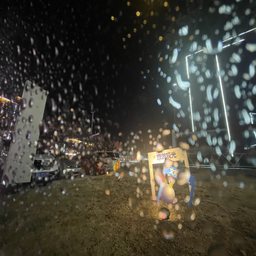}}\hfill	
		\subfloat{\includegraphics[width=0.164\textwidth]{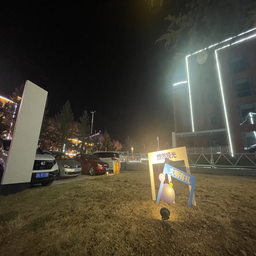}}\hfill
		\subfloat{\includegraphics[width=0.164\textwidth]{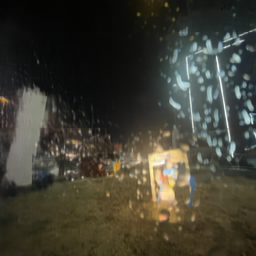}}\hfill
		\subfloat{\includegraphics[width=0.164\textwidth]{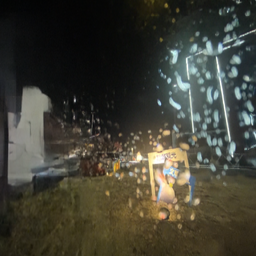}}\hfill
		\subfloat{\includegraphics[width=0.164\textwidth]{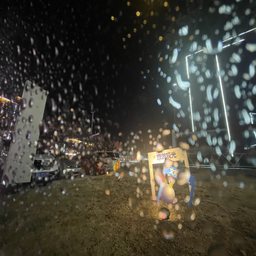}}\hfill
		\subfloat{\includegraphics[width=0.164\textwidth]{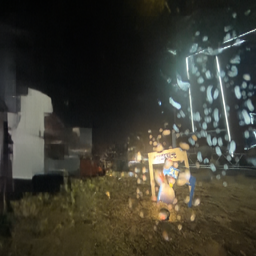}}\hfill
		\setcounter{subfigure}{0}
		\subfloat[Input]{\includegraphics[width=0.164\textwidth]{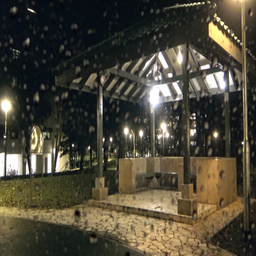}}\hfill	
		\subfloat[Ground Truth]{\includegraphics[width=0.164\textwidth]{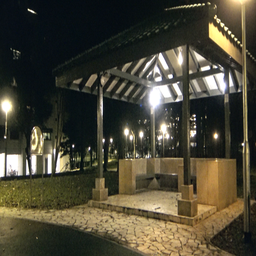}}\hfill
		\subfloat[IDT~\cite{xiao2022image}]{\includegraphics[width=0.164\textwidth]{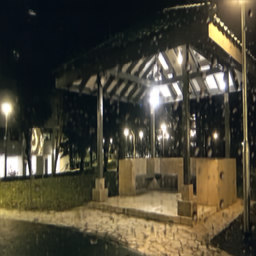}}\hfill
		\subfloat[RDiffusion~\cite{ozdenizci2023restoring}]{\includegraphics[width=0.164\textwidth]{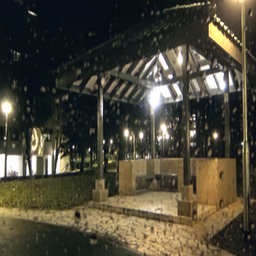}}\hfill
		\subfloat[Restormer~\cite{Restormer}]{\includegraphics[width=0.164\textwidth]{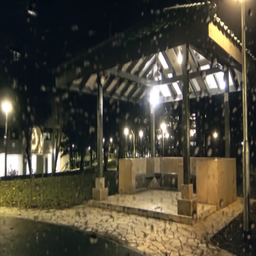}}\hfill
		\subfloat[DiT~\cite{peebles2022scalable}]{\includegraphics[width=0.164\textwidth]{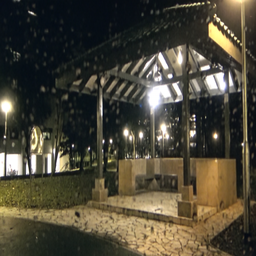}}\hfill
	\caption{Comparison of methods on nighttime background-focused images, covering both raindrop algorithms and restoration backbones.}
	\label{fig:nightbd}
\end{figure*} 

{\flushleft \bf Comparison with Existing Datasets.}
The difference between our Raindrop Clarity dataset and the existing datasets is shown in Table.~\ref{tb:compare} and Fig.~\ref{fig:dataset}.

The stereo RobotCar dataset \cite{porav2019can} includes 4,818 image pairs, randomly selected for training, validation, and testing.
However, the raindrops in this dataset are disproportionately large and lack realism.

The raindrop dataset proposed by Qian~\etal~\cite{qian2018attentive} consists of 861 training pairs and 58 testing pairs. 
While it features common raindrop shapes, it lacks the diversity of real-world scenes~\cite{you2013adherent}.
Also, the backgrounds between the raindrop images and the ground truth are not perfectly aligned~\cite{wen2023video}.

The RainDS-Real dataset, proposed by Quan~\etal~\cite{quan2021removing}, comprises 248 quadruplets, including raindrops, rain streaks, and their corresponding ground truths.
The RainDS-Synthetic dataset has a domain gap with the real raindrop images. 
Moreover, the real-world raindrops are consistently small, obscuring only a fraction of the image details, making recovery less challenging.

The Windshield dataset, proposed by Soboleva and Shipitko~\cite{soboleva2021raindrops}, was captured by a camera attached to the vehicle during its movement in urban areas and highway environments.
The Windshield does not provide the corresponding clean pairs, making this dataset not available for robust raindrop removal method testing.
Additionally, frequent moving objects further complicate the dataset. 

Existing raindrop datasets, though valuable, have certain limitations. 
They do not include raindrop-focused and nighttime raindrops.
In addition, we have noticed reflection errors in the datasets from Qian~\cite{qian2018attentive} and RainDS-Real~\cite{quan2021removing}.

\begin{figure}[t!]
	\centering
	\captionsetup[subfig]{labelformat=empty}
	\setcounter{subfigure}{0}
	{\includegraphics[width=\linewidth]{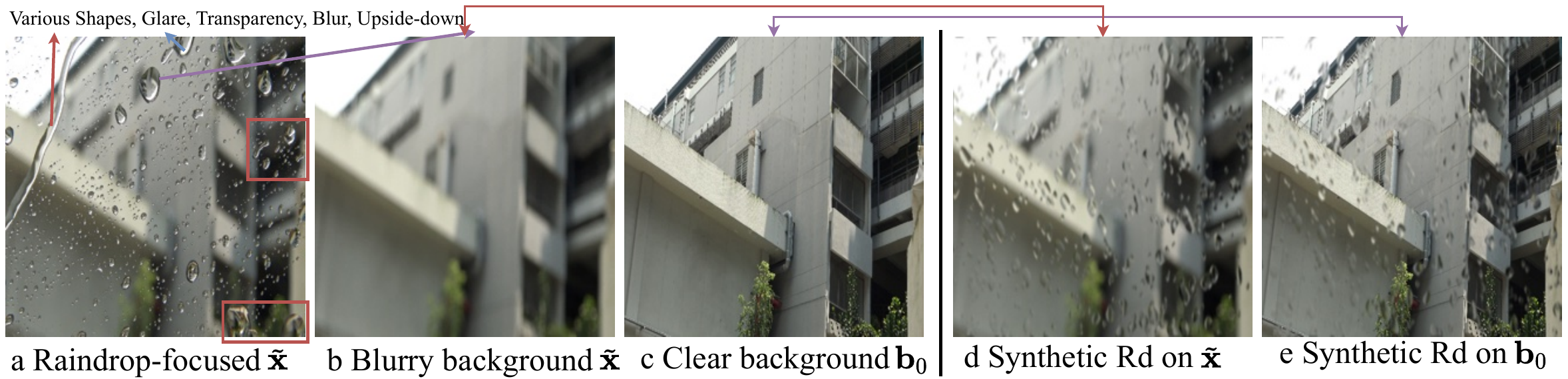}}\hfill
	\caption{We show the differences between the synthetic raindrops~\cite{pizzati2023physics} and real-world raindrops from Raindrop Clarity.}
	\label{fig:syn}
\end{figure}

{\flushleft \bf Comparison with Synthetic Raindrops.}
In Fig.~\ref{fig:syn}, we show the existing state-of-the-art synthetic physical model~\cite{pizzati2023physics}, although it can generate raindrops on blurry and clear backgrounds (Fig.~\ref{fig:syn} d generated on b, Fig.~\ref{fig:syn} e generated on c). It still has a gap with real-world collected raindrops.
Real-world collected raindrops display diverse appearances, including various shapes such as round, elliptical, and long strips.
Also, the transparency and glare effects of raindrops collected in the real-world varies.

{\flushleft \bf Quantitative Comparison.}
To effectively evaluate the results on our proposed datasets, we use three popular metrics PSNR, SSIM~\cite{ssim}, and LPIPS~\cite{lpips} in generation tasks~\cite{jin2021dc,jin2022unsupervised,jin2023enhancing,jin2024des3,ye2022perceiving,chen2023sparse,chen2023msp,ye2023adverse,chen2022snowformer,wang2024restoring,wang2023seeing,li2024sed,li2023learning,li2023diffusion,li2020learning,luo2023crefnet,luo2024intrinsicdiffusion,wang2024selfpromer,wang2024correlation,wang2023promptrestorer}. 

The qualitative results are shown in Table \ref{tab:PSNR_SSIM}. 
For the daytime dataset, the PSNR and SSIM of the original input images are 21.92 and 0.560. 
Raindrop algorithm IDT~\cite{xiao2022image} achieves the best performance (26.05 PSNR of and 0.736 of SSIM) compared with other raindrop algorithms. 
For the restoration backbones, DiT~\cite{peebles2022scalable} achieves a PSNR of 26.03 and a SSIM of 0.752.
Restormer~\cite{Restormer} achieves a PSNR of 26.08 and a SSIM of 0.748, which also performs competitively.

It can be observed that, for the nighttime dataset, PSNR and SSIM of the input images are 24.78 and 0.726. 
Among the raindrop algorithms, IDT~\cite{xiao2022image} achieves better performance than other methods, with a PSNR of 26.81 and a SSIM of 0.851.
For the restoration backbones,  Restormer~\cite{Restormer} achieves the highest PSNR of 26.87, while DiT~\cite{peebles2022scalable} achieves the lowest LPIPS of 0.111, which outperforms other backbones.

{\flushleft \bf Qualitative Comparison.}
Figs.~\ref{fig:dayrd}-\ref{fig:nightbd} show the qualitative results of state-of-the-art methods. 
To be specific, Fig.~\ref{fig:dayrd} and Fig.~\ref{fig:daybd} show the results of daytime raindrop removal. 
The input images in Fig.~\ref{fig:dayrd} and Fig.~\ref{fig:daybd} are daytime raindrop-focused and background-focused, respectively. 
Fig.~\ref{fig:nightrd} and Fig.~\ref{fig:nightbd} show the results of nighttime raindrop-focused and background-focused removal, respectively.

It can be found that, for daytime and nighttime raindrop-focused removal in Fig.~\ref{fig:dayrd} and Fig.~\ref{fig:nightrd}, existing state-of-the-art methods fail to remove raindrops and recover the background. 
This is because most of them are designed for background-focused scenes and thus neglect the blurry problem. 
To our knowledge, Restormer~\cite{Restormer} and Uformer~\cite{Uformer} are specifically designed to address deblurring problem.
However, they still struggle with recovering details and textures.
As a result, these methods are struggling to simultaneously remove raindrops and recover the background.

In Fig.~\ref{fig:daybd}, for daytime background-focused inputs, we can observe that existing state-of-the-art methods can handle most raindrops. However, they are struggling to address long strips of raindrops. This is because their methods may not detect long strips of raindrops and thus fail to remove them. 
Furthermore, in Fig.~\ref{fig:nightbd}, we find that existing state-of-the-art methods are struggling to handle nighttime background-focused samples. To be specific, raindrops in these samples cannot be totally removed and details of the restored images are not clear. This is because nighttime raindrops usually reflect artificial light and thus are more complex than daytime raindrops. 
Meanwhile, existing state-of-the-art methods are designed for daytime raindrop removal and neglect the effect of complex nighttime conditions. Thus, existing state-of-the-art methods cannot achieve promising results.

\section{Conclusion}
In this paper, we propose a new raindrop dataset, named Raindrop Clarity. It includes a total of 15,186 high-quality pairs/triplets of images with raindrops and the corresponding clear background images. Moreover, our dataset is collected under various conditions, including different scenes, times (daytime and nighttime), and focuses (raindrop- and background-focused). 
To be specific, 3,606 daytime raindrop-focused, 1,836 daytime background-focused, 4,838 nighttime raindrop-focused and 4,906 nighttime background-focused images are collected.
With the proposed dataset, we re-analyze existing state-of-the-art methods, including raindrop algorithms and restoration backbones, and point out several open problems: raindrop-focus raindrop removal and nighttime raindrop removal.
We wish to draw the attention of the community to the open problems.

\section{Social Impacts}
Existing raindrop datasets focus on daytime background-focused raindrop removal. However,  in the real world, raindrop images may focus on raindrops, resulting in a blurry background. 
Also, raindrops may occur in nighttime images.
To draw the attention of the community to the open problems, we propose a new raindrop dataset, named Raindrop Clarity.
It mainly collected from four conditions: daytime raindrop-focused, daytime background-focused, nighttime raindrop-focused and nighttime background-focused.
Experimental results on our proposed dataset show that existing state-of-the-art methods cannot effectively
remove raindrops under complex conditions. 
As a result, we believe that further exploration of our Raindrop Clarity dataset can help the community to address more complex raindrops.

\section*{Acknowledgments}
This research is supported by the National Research Foundation, Singapore, under its AI Singapore Programme (AISG Award No: AISG2-PhD/2022-01-037[T]).
Xin Li's research is supported by NSFC under Grant 623B2098.	
Malu Zhang's work is supported by the National Science Foundation of China under Grant 62106038, and in part by the Sichuan Science and Technology Program under Grant 2023YFG0259.
We thank Robby T. Tan and Beibei Lin for their valuable discussions, revisions, and feedback.

\bibliographystyle{splncs04}
\bibliography{main}
\end{document}